\RequirePackage{fix-cm}
\documentclass[11pt]{article}
\usepackage[preprint]{acl}
\usepackage{times}
\usepackage{latexsym}
\usepackage[T1]{fontenc}
\usepackage[utf8]{inputenc}
\usepackage{microtype}
\usepackage{amsmath,amssymb}
\usepackage{booktabs,tabularx,array,multirow}
\usepackage{graphicx}
\usepackage{xcolor}
\usepackage{tikz}
\usepackage{cuted}
\usepackage{caption}
\usepackage{booktabs}
\usepackage{caption}
\usepackage{graphicx}
\usetikzlibrary{arrows.meta,positioning,calc,fit,backgrounds}
\usepackage{enumitem}
\usepackage{placeins}
\usepackage{balance}
\definecolor{ink}{HTML}{243347}
\definecolor{blueaccent}{HTML}{2867AB}
\definecolor{bluewash}{HTML}{EDF4FC}
\definecolor{tealaccent}{HTML}{087F83}
\definecolor{tealwash}{HTML}{E9F6F3}
\definecolor{violetaccent}{HTML}{7250A2}
\definecolor{violetwash}{HTML}{F3EFF9}
\definecolor{amberaccent}{HTML}{A96A13}
\definecolor{amberwash}{HTML}{FFF5E1}
\definecolor{roseaccent}{HTML}{AF4A63}
\definecolor{rosewash}{HTML}{FCEDF1}
\definecolor{papergray}{HTML}{F7F9FB}
\definecolor{rulegray}{HTML}{CCD5DF}
\definecolor{figNavy}{HTML}{16324F}
\definecolor{figBlue}{HTML}{2563EB}
\definecolor{figIndigo}{HTML}{5B47CD}
\definecolor{figTeal}{HTML}{00897B}
\definecolor{figAmber}{HTML}{CB8500}
\definecolor{figPink}{HTML}{C43D83}
\definecolor{figSlate}{HTML}{52677D}
\definecolor{figBorder}{HTML}{D6E0EB}

\newcommand{\mindset}{MINDSET}
\newcommand{\act}[1]{\textsc{#1}}
\newcommand{\best}[1]{\textbf{#1}}
\newcolumntype{Y}{>{\raggedright\arraybackslash}X}
\title{MINDSET: Energy-based Memory Evolution for Long \\ Conversational Agents}
\author{\textbf{Sujato Dutta} \quad \textbf{Sreekruthy Tummala} \quad \textbf{Shashank Vanga} \\ \quad \textbf{Ayushmi Pavani} \quad
        \textbf{Neeraj Choudhary} \\
        Mahindra University}
\begin{document}
\maketitle
\begin{abstract}
In conversational memory, a later statement may replace an earlier fact, describe a different context, or be an isolated contradiction. Thus, agentic memory must track what changed over time without erasing what was true before. We introduce \mindset, a memory controller that makes this distinction through explicit state transitions. It preserves original dialogue episodes and organizes them into versioned schemas. For each new episode, it chooses whether to reinforce a schema, supersede its current state, create a contextual branch, or start a new schema. A local energy score balances coherence, contradiction, historical preservation and fragmentation. Stability rules prevent isolated conflicts from immediately replacing established memory. On a frozen 700-question LoCoMo subset, \mindset\ achieves the highest observed answer F1 among six evaluated systems and significantly improves retrieval ranking over LightMem. Without retuning, it obtains 69.23\% judge accuracy on all LongMemEval knowledge-update cases, compared with 64.10\% for Vector RAG. Two additional answer model backbones using Gemma-4-31B and GLM-4.7 provide further qualified evidence of generalization. These findings support explicit memory-state management as a useful approach to changing conversational knowledge.
\end{abstract}

\section{Introduction}
\label{sec:introduction}
A conversational assistant may learn that a user lives in Delhi and later that the user has moved to Mumbai. Answering ``Where does the user live now?'' requires the new fact; answering ``Where did the user live before moving?'' requires the old one. A different statement, such as ``I stay in Delhi when visiting family,'' may describe a coexisting context rather than a correction. A useful memory system must distinguish these cases while preserving the evidence behind each answer.

Storing every turn in a vector index retains the record but leaves the answer model to resolve changing and conflicting facts repeatedly. Replacing earlier memories with a summary can reduce context, but may obscure when a fact applied or why it was revised. Full-history prompting is another option, although longer contexts do not guarantee reliable evidence use \citep{ref28}. The problem is therefore not only retrieving a relevant statement. It is deciding how the memory state should evolve when a new statement arrives.

Prior work addresses memory updates, temporal validity, and provenance. \citet{ref24} study outdated information in long-term conversation; Mem0 extracts and consolidates memories \citep{ref38}; Zep/Graphiti maintains temporally grounded relationships \citep{rasmussen2025zep}; and APEX-MEM preserves interaction history in an append-only temporal graph \citep{banerjee2026apex}. Within this direction, \mindset's contribution is \emph{explicit selection among competing memory-state transitions through a common local energy score, stability rules and provenance-preserving updates}.

\mindset\ separates the original conversational record from its organization. Raw episodes remain unchanged. A schema indexes episodes that support a compatible state, records exceptions, and links earlier and later versions. Each incoming episode can reinforce the current schema, supersede it, split off a contextual branch, or create an independent schema. The controller compares the costs of merging incompatible evidence and creating unnecessary structure. At question time, retrieval uses these states as routes back to original episodes. The answer model receives a bounded evidence packet rather than a rewritten account of the conversation.

We examine this design through three questions. First, does it improve evidence retrieval and answering over established memory baselines? Second, does a controller calibrated on LoCoMo transfer to an explicitly update-focused benchmark without retuning? Third, which design choices contribute to its behavior, and what efficiency costs do they impose? We use a frozen LoCoMo-700 evaluation, all 78 LongMemEval knowledge-update cases, six ablations and additional answer model backbones.

Our contributions are: (1) an explicit, energy-based controller for conversational memory transitions; (2) a provenance-preserving representation that combines stable states with accessible historical evidence; and (3) a controlled evaluation of retrieval quality, answering, and computational cost.

\section{Related Work}
\label{sec:related}
\paragraph{Retrieval and conversational memory.}
Retrieval-augmented generation supplies language models with external evidence \citep{ref7}. Sentence embeddings support efficient semantic matching \citep{ref9}, but retrieval alone does not determine whether a fact remains current. MemoryBank combines dialogue memories, summaries, and a time-dependent updating mechanism \citep{ref4}. LightMem separates short-, mid-, and long-term storage, with bounded online retrieval and offline consolidation \citep{ref5}. Nemori organizes interactions into episodes and distills knowledge through prediction error \citep{ref6}. \mindset\ instead emphasizes the explicit choice of state transition when incorporating a new episode.

\paragraph{Updating and temporal organization.}
Temporal memory is handled in many different ways.  \citet{ref24} manage invalidated and redundant memories. THEANINE preserves the contextual value of outdated memories through timeline-based management \citep{ref17}, while RMM combines prospective and retrospective reflection \citep{ref16}. Mem0 dynamically extracts, consolidates, and retrieves information \citep{ref38}. Zep uses Graphiti to maintain temporal knowledge graphs and historical relationships \citep{rasmussen2025zep}. In contrast, \mindset\ explicitly compares reinforcement, supersession, contextual branching and new-state creation during ingestion. This distinction concerns the control rule.

\paragraph{Broader memory architectures and evaluation.}
MIRIX coordinates multiple agents and memory types \citep{wang2025mirix}; MemOS treats memory as an operational resource spanning representations and lifecycle management \citep{li2025memos}. These are broader architectures than our transition controller and may offer complementary capabilities. LoCoMo tests reasoning over long conversations \citep{ref1}; LongMemEval explicitly includes knowledge updates \citep{ref3}. Our experiments evaluate selected abilities under a shared local protocol rather than establishing a ranking across published leaderboards.

\section{MINDSET}
\label{sec:method}
\begin{figure*}[t]
  \centering
  \resizebox{\textwidth}{!}{
\begin{tikzpicture}[x=1cm,y=1cm,text=figNavy,
 lab/.style={font=\sffamily\fontsize{7.4}{9}\selectfont,inner sep=0pt,align=left},
 hd/.style={lab,font=\sffamily\bfseries\fontsize{9}{10.5}\selectfont},
 mini/.style={lab,font=\sffamily\fontsize{6.5}{8}\selectfont},
 box/.style={rounded corners=5pt,line width=.65pt},
 flow/.style={-{Stealth[length=1.9mm,width=1.4mm]},draw=figNavy!75,line width=.85pt},
 ico/.style={line width=.85pt,line cap=round,line join=round}]
\path[use as bounding box] (-.15,-.07) rectangle (16.22,7.6);
\fill[figNavy] (0,7.34) circle (.14);
\node[lab,text=white,font=\sffamily\bfseries\fontsize{7}{8}\selectfont] at (0,7.34) {A};
\node[hd,anchor=west] at (.28,7.34) {INGEST \& EVOLVE};
\node[lab,anchor=east,text=figSlate] at (16.2,7.34) {Local control\enspace$\cdot$\enspace No generative rewriting};
\foreach \xl/\xr/\col in {0/3.4/figBlue,4/10.9/figIndigo,11.7/16.2/figTeal}{
 \draw[box,draw=\col!35,fill=\col!3] (\xl,2.72) rectangle (\xr,6.96);
 \begin{scope}
  \clip[rounded corners=5pt] (\xl,2.72) rectangle (\xr,6.96);
  \fill[figNavy] (\xl,6.14) rectangle (\xr,6.96);
 \end{scope}
}
\node[hd,text=white,anchor=west] at (.22,6.55) {01\quad Episodes};
\node[hd,text=white,anchor=west] at (4.25,6.55) {02\quad Transition controller};
\node[hd,text=white,anchor=west] at (11.94,6.55) {03\quad Versioned schemas};
\draw[box,draw=figBlue!25,fill=figBlue!12] (.69,4.35) rectangle (3.07,5.92);
\draw[box,draw=figBlue!35,fill=figBlue!6] (.53,4.21) rectangle (2.91,5.78);
\draw[box,draw=figBlue!60,fill=white] (.37,4.07) rectangle (2.75,5.64);
\fill[figBlue!11] (.38,5.14) rectangle (2.74,5.5);
\node[lab,font=\sffamily\bfseries\fontsize{7.5}{9}\selectfont,anchor=west] at (.54,5.31) {Original dialogue};
\foreach \y/\xr in {4.83/2.45,4.52/2.13}{
 \fill[figBlue!80] (.6,\y) circle (.045);
 \draw[draw=figSlate!55,line width=.8pt] (.8,\y) -- (\xr,\y);
}
\node[mini,anchor=west,text=figSlate] at (.54,4.25) {time\enspace$\cdot$\enspace speaker};
\draw[flow,draw=figBlue!70] (1.7,3.98) -- (1.7,3.73);
\draw[box,draw=figBlue!45,fill=figBlue!10] (.3,3.17) rectangle (3.1,3.67);
\node[lab,font=\sffamily\bfseries\fontsize{7.5}{9}\selectfont,text=figNavy] at (1.7,3.42) {Append-only raw store};
\node[mini,text=figSlate] at (1.7,2.93) {Raw text stays unchanged};
\node[lab,text=figSlate] at (7.45,5.85) {Feasibility\quad$\rightarrow$\quad local energy\quad$\rightarrow$\quad stability};
\foreach \xl/\xr/\yb/\yt/\col in {4.23/7.31/4.32/5.52/figTeal,7.59/10.67/4.32/5.52/figBlue,4.23/7.31/2.98/4.18/figAmber,7.59/10.67/2.98/4.18/figPink}{
 \draw[box,draw=\col!50,fill=\col!10] (\xl,\yb) rectangle (\xr,\yt);
 \fill[\col,rounded corners=1.5pt] (\xl,\yb+.15) rectangle (\xl+.065,\yt-.15);
}
\foreach \x/\y/\col in {4.62/5.12/figTeal,7.98/5.12/figBlue,4.62/3.78/figAmber,7.98/3.78/figPink}
 \fill[\col] (\x,\y) circle (.18);
\draw[ico,draw=white] (4.53,5.12) -- (4.71,5.12) (4.62,5.03) -- (4.62,5.21);
\draw[ico,draw=white,-{Stealth[length=1.1mm,width=.85mm]}] (7.87,5.12) -- (8.09,5.12);
\draw[ico,draw=white] (4.53,3.78) -- (4.61,3.78) -- (4.71,3.88) (4.61,3.78) -- (4.71,3.68);
\draw[ico,draw=white] (7.9,3.7) rectangle (8.06,3.86);
\node[hd,anchor=west,text=figTeal,font=\sffamily\bfseries\fontsize{8.3}{10}\selectfont] at (4.94,5.12) {REINFORCE};
\node[hd,anchor=west,text=figBlue,font=\sffamily\bfseries\fontsize{8.3}{10}\selectfont] at (8.3,5.12) {SUPERSEDE};
\node[hd,anchor=west,text=figAmber!85!black,font=\sffamily\bfseries\fontsize{8.3}{10}\selectfont] at (4.94,3.78) {SPLIT};
\node[hd,anchor=west,text=figPink,font=\sffamily\bfseries\fontsize{8.3}{10}\selectfont] at (8.3,3.78) {NEW};
\node[lab,anchor=west] at (4.49,4.65) {Support / exception};
\node[lab,anchor=west] at (7.85,4.65) {Link state versions};
\node[lab,anchor=west] at (4.49,3.31) {Branch by context};
\node[lab,anchor=west] at (7.85,3.31) {Independent schema};
\draw[box,draw=figBlue!50,fill=figBlue!10] (11.98,4.95) rectangle (13.68,5.78);
\draw[box,draw=figTeal!55,fill=figTeal!12] (14.2,4.95) rectangle (15.92,5.78);
\node[lab,font=\sffamily\bfseries\fontsize{7.5}{9}\selectfont,text=figBlue] at (12.83,5.49) {Prior state};
\node[mini,text=figSlate] at (12.83,5.17) {historical};
\node[lab,font=\sffamily\bfseries\fontsize{7.5}{9}\selectfont,text=figTeal] at (15.06,5.49) {Current state};
\node[mini,text=figSlate] at (15.06,5.17) {active};
\draw[flow,draw=figBlue,line width=.7pt] (13.68,5.37) -- (14.2,5.37);
\draw[box,draw=figPink!50,fill=figPink!10] (11.98,3.89) rectangle (13.68,4.53);
\draw[box,draw=figAmber!60,fill=figAmber!13] (14.2,3.89) rectangle (15.92,4.53);
\node[lab,font=\sffamily\bfseries\fontsize{7}{8.5}\selectfont,text=figPink] at (12.83,4.21) {New state};
\node[lab,font=\sffamily\bfseries\fontsize{7}{8.5}\selectfont,text=figAmber!85!black] at (15.06,4.21) {Contextual};
\draw[draw=figSlate!65,densely dashed,line width=.65pt] (12.15,4.95) -- (11.87,4.95) -- (11.87,3.57) -- (16.03,3.57) -- (16.03,4.95) -- (15.72,4.95);
\foreach \x in {13.5,14.4} \draw[draw=figSlate!65,densely dashed,line width=.65pt] (\x,3.89) -- (\x,3.57);
\foreach \x/\col in {13.1/figBlue,13.95/figPink,14.81/figAmber}{
 \draw[rounded corners=1.5pt,draw=\col!60,fill=white,line width=.6pt] (\x-.24,3.18) rectangle (\x+.24,3.55);
 \draw[draw=\col!60,line width=.5pt] (\x-.14,3.42) -- (\x+.14,3.42) (\x-.14,3.3) -- (\x+.08,3.3);
}
\node[mini,text=figSlate] at (13.95,2.94) {Provenance to original episodes};
\draw[flow] (3.4,4.83) -- (4,4.83);
\draw[flow] (10.9,4.83) -- (11.7,4.83);
\fill[figTeal] (0,2.29) circle (.14);
\node[lab,text=white,font=\sffamily\bfseries\fontsize{7}{8}\selectfont] at (0,2.29) {B};
\node[hd,anchor=west,text=figTeal] at (.28,2.29) {RETRIEVE \& ANSWER};
\draw[-{Stealth[length=1.65mm]},draw=figTeal!80,densely dashed,line width=.75pt] (13.95,2.72) -- (13.95,2.1) -- (7.45,2.1) -- (7.45,1.73);
\node[mini,anchor=south,text=figTeal] at (10.85,2.19) {follow provenance};
\draw[box,draw=figBlue!35,fill=figBlue!5] (0,.16) rectangle (3.4,1.73);
\fill[figBlue!12] (.3,1.14) circle (.18);
\node[lab,text=figNavy,font=\sffamily\bfseries\fontsize{8}{9}\selectfont] at (.3,1.14) {?};
\node[hd,anchor=west] at (.65,1.3) {Query intent};
\node[lab,anchor=west,text=figSlate] at (.65,.9) {Current / historical};
\node[lab,anchor=west,text=figSlate] at (.65,.5) {Contextual / generic};
\foreach \xl/\xr/\col in {4/6.04/figTeal,6.43/8.47/figIndigo,8.86/10.9/figPink}{
 \draw[box,draw=\col!45,fill=\col!8] (\xl,.16) rectangle (\xr,1.73);
 \fill[\col,rounded corners=1.5pt] (\xl+.16,1.56) rectangle (\xr-.16,1.6);
}
\draw[ico,draw=figTeal] (4.95,1.21) circle (.145);
\draw[ico,draw=figTeal] (5.05,1.1) -- (5.19,.97);
\draw[ico,draw=figIndigo] (7.45,1.33) -- (7.24,1.02) (7.45,1.33) -- (7.66,1.02);
\foreach \x/\y in {7.45/1.33,7.24/1.02,7.66/1.02} \fill[figIndigo] (\x,\y) circle (.055);
\draw[ico,draw=figPink] (9.66,1.35) -- (10.08,1.35) (9.66,1.18) -- (9.99,1.18) (9.66,1.01) -- (9.89,1.01);
\node[hd,text=figTeal,font=\sffamily\bfseries\fontsize{8}{9.5}\selectfont] at (5.02,.76) {Search};
\node[hd,text=figIndigo,font=\sffamily\bfseries\fontsize{8}{9.5}\selectfont] at (7.45,.76) {Expand};
\node[hd,text=figPink,font=\sffamily\bfseries\fontsize{8}{9.5}\selectfont] at (9.88,.76) {Rerank};
\node[mini,text=figSlate] at (5.02,.39) {Raw + schemas};
\node[mini,text=figSlate] at (7.45,.39) {Original evidence};
\node[mini,text=figSlate] at (9.88,.39) {Select evidence};
\draw[flow,line width=.75pt] (3.4,.95) -- (4,.95);
\draw[flow,line width=.75pt] (6.04,.95) -- (6.43,.95);
\draw[flow,line width=.75pt] (8.47,.95) -- (8.86,.95);
\draw[flow,line width=.75pt] (10.9,.95) -- (11.7,.95);
\draw[box,draw=figTeal!45,fill=figTeal!6] (11.7,.16) rectangle (16.2,1.73);
\node[hd,anchor=west,text=figTeal] at (11.94,1.37) {Bounded evidence packet};
\node[lab,anchor=west,text=figSlate] at (11.94,.97) {Raw evidence + source IDs};
\fill[figNavy,rounded corners=3pt] (11.94,.35) rectangle (15.96,.71);
\node[lab,text=white,font=\sffamily\bfseries\fontsize{7.2}{8.5}\selectfont] at (13.95,.53) {Frozen answer model};
\end{tikzpicture}}
  \caption{\textbf{MINDSET overview.} During ingestion, an immutable dialogue episode is assigned through an explicit transition controller. Schemas retain active states, historical versions, and contextual branches, each linked to original evidence. During answering, query intent guides retrieval through raw episodes and schema versions}
  \label{fig:overview}
\end{figure*}
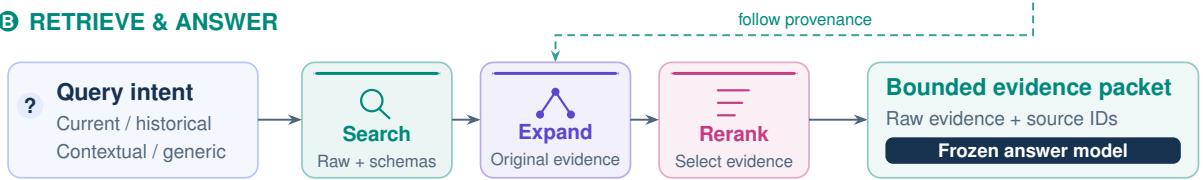

Figure~\ref{fig:overview} separates memory construction from question answering. This section first defines the stored objects, then describes how the controller updates and retrieves them.

\subsection{Episodes and Versioned Schemas}
\label{sec:objects}
An \emph{episode} is an original dialogue turn with its conversation ID, session, timestamp, speaker, and text. Local feature extraction adds entities, roles, contexts, attributes, and lexical change cues. These features support indexing and transition decisions; they do not replace the original text.

A \emph{schema} is a versioned state over compatible episodes. It stores supporting episode IDs, exception IDs, embedding statistics, validity bounds, active or historical status, and predecessor/successor links. Supporting episodes agree sufficiently with the state. Exceptions retain conflicting observations that have not justified a structural change. A schema is thus an index over evidence, rather than a free-form summary presented as a new source. Superseding a schema closes its active version and creates a linked successor. A contextual branch creates a coexisting active schema, allowing different states to remain available in different settings. In every case, the raw episodes remain available. This distinction lets the system change its interpretation of the record without overwriting the record itself.

\subsection{Four Possible Transitions}
\label{sec:actions}
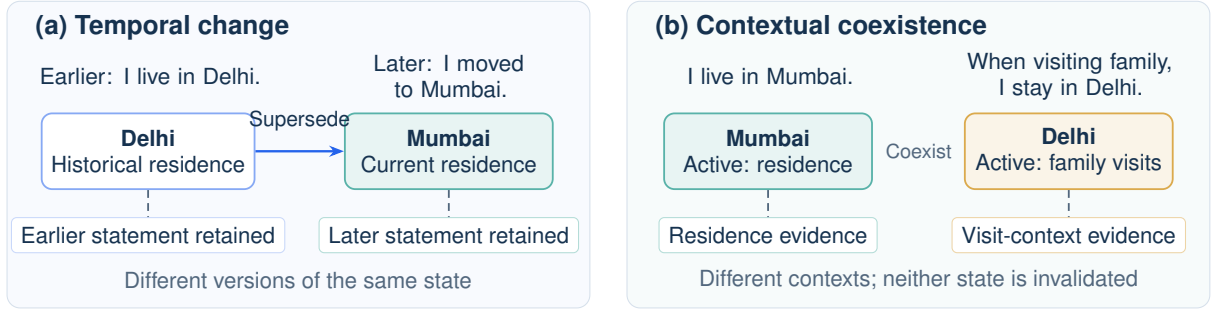
\begin{figure*}[t]
  \centering
  \resizebox{\textwidth}{!}{
\begin{tikzpicture}[x=1cm,y=1cm,
  text=figNavy,
  label/.style={font=\sffamily\fontsize{8.5}{10}\selectfont,align=center},
  heading/.style={label,font=\sffamily\bfseries\fontsize{10}{12}\selectfont},
  state/.style={label,rounded corners=4pt,line width=.7pt,
    minimum width=2.8cm,minimum height=1.02cm},
  flow/.style={-{Stealth[length=2mm]},line width=.9pt,draw=figBlue}]
\path[use as bounding box] (0,0) rectangle (16.2,4.18);
\draw[rounded corners=5pt,draw=figBorder,fill=figBlue!3]
  (0,.02) rectangle (7.85,4.15);
\draw[rounded corners=5pt,draw=figBorder,fill=figTeal!3]
  (8.35,.02) rectangle (16.2,4.15);
\node[heading,anchor=west] at (.25,3.8) {(a) Temporal change};
\node[heading,anchor=west] at (8.6,3.8) {(b) Contextual coexistence};
\node[label,text width=3cm] at (1.9,3.14)
  {Earlier: I live in Delhi.};
\node[label,text width=3cm] at (5.95,3.14)
  {Later: I moved to Mumbai.};
\node[state,draw=figBlue!55,fill=white] (old) at (1.9,2.13)
  {\textbf{Delhi}\\Historical residence};
\node[state,draw=figTeal!65,fill=figTeal!9] (current) at (5.95,2.13)
  {\textbf{Mumbai}\\Current residence};
\draw[flow] (old.east) -- node[label,above=4pt,font=\sffamily\fontsize{8}{9.5}\selectfont]
  {Supersede} (current.west);
\draw[densely dashed,draw=figSlate,line width=.7pt] (old.south) -- (1.9,1.22);
\draw[densely dashed,draw=figSlate,line width=.7pt] (current.south) -- (5.95,1.22);
\node[label,draw=figBlue!25,fill=white,rounded corners=2pt,
  minimum width=2.8cm,minimum height=.44cm] at (1.9,1)
  {Earlier statement retained};
\node[label,draw=figTeal!30,fill=white,rounded corners=2pt,
  minimum width=2.8cm,minimum height=.44cm] at (5.95,1)
  {Later statement retained};
\node[label,text=figSlate] at (3.92,.4)
  {Different versions of the same state};
\node[label,text width=3.1cm] at (10.25,3.14)
  {I live in Mumbai.};
\node[label,text width=3.1cm] at (14.3,3.14)
  {When visiting family,\\I stay in Delhi.};
\node[state,draw=figTeal!65,fill=figTeal!9] (home) at (10.25,2.13)
  {\textbf{Mumbai}\\Active: residence};
\node[state,draw=figAmber!70,fill=figAmber!9] (visit) at (14.3,2.13)
  {\textbf{Delhi}\\Active: family visits};
\node[label,text=figSlate,font=\sffamily\fontsize{7.5}{9}\selectfont,
  text width=1.1cm] at (12.28,2.13) {Coexist};
\draw[densely dashed,draw=figSlate,line width=.7pt] (home.south) -- (10.25,1.22);
\draw[densely dashed,draw=figSlate,line width=.7pt] (visit.south) -- (14.3,1.22);
\node[label,draw=figTeal!30,fill=white,rounded corners=2pt,
  minimum width=2.8cm,minimum height=.44cm] at (10.25,1)
  {Residence evidence};
\node[label,draw=figAmber!35,fill=white,rounded corners=2pt,
  minimum width=2.8cm,minimum height=.44cm] at (14.3,1)
  {Visit-context evidence};
\node[label,text=figSlate] at (12.28,.4)
  {Different contexts; neither state is invalidated};
\end{tikzpicture}}
  \caption{\textbf{Change is not coexistence.} Left: a later move creates a successor and preserves the earlier residence as historical. Right: different settings can remain active together without a supersession link. Both retain access to the original statements.}
  \label{fig:state-evolution}
\end{figure*}

The controller retrieves nearby active schemas and selects a target using compatibility signals. It then considers four actions:
\begin{itemize}[leftmargin=*,itemsep=2pt,topsep=3pt]
  \item \act{Reinforce}: retain the schema and attach the episode as supporting evidence.
  \item \act{Supersede}: mark the current version historical and create a linked successor representing a later state.
  \item \act{Split}: create a coexisting contextual branch while preserving the earlier state.
  \item \act{New}: start an independent schema when the episode does not fit the available states.
\end{itemize}
Figure~\ref{fig:state-evolution} illustrates the distinction between supersession and contextual coexistence. Supersession preserves Delhi as a historical state while making Mumbai current; visiting Delhi in a different context need not invalidate the current residence.

\subsection{Compatibility, Change, and Context}
\label{sec:signals}
Compatibility combines semantic similarity with entity, role, context, and topic matches. Entity, role, and context matches measure how much the incoming episode overlaps the schema's accumulated feature counts. The topic signal compares the extracted attribute slot with the schema topic. These checks reduce the risk of merging semantically similar statements about different people or situations.

Contradiction is assessed for attributes treated as single-valued. When an incoming value conflicts with values already stored for the same slot, it raises the cost of merging the episode into the current state. The comparison uses normalized lexical overlap rather than a general-purpose entailment model. Multi-valued attributes, including preferences, plans, possessions, experiences, abilities, and dislikes, are excluded from this contradiction test.

Temporal evidence combines a conflict with chronology, shared context, explicit change cues, and recurrence among stored exceptions. Contextual evidence instead favors branching when the episode concerns a related topic but differs in context, role, or semantic fit. These signals first restrict which actions are feasible: a detected conflict is required for supersession, contextual evidence is required for splitting, and a strong mismatch can trigger a new schema. Energy comparison occurs only after these checks.

\subsection{Choosing a Transition}
\label{sec:energy}
For each feasible action, the controller computes a local energy:
\begin{equation}
\label{eq:energy}
 \begin{aligned}
 E(a)={}&w_D D(a)+w_C C(a)+w_H H(a)\\
       &+w_F F(a)+w_I I(a).
 \end{aligned}
\end{equation}
Here, \emph{distortion} $D$ measures the representational cost of absorbing the episode; \emph{contradiction} $C$ penalizes unresolved single-valued conflicts; \emph{historical damage} $H$ penalizes shifting an established state to accommodate incompatible evidence; \emph{fragmentation} $F$ discourages unnecessary and weakly supported schemas; and \emph{inconsistency} $I$ measures heterogeneous schema features and disagreement between the action and temporal/contextual signals.

These costs encode a practical trade-off. Always merging produces incoherent states; always creating new states produces fragmented memory. Supersession and splitting preserve the earlier state, but should be used only when the evidence supports a temporal change or contextual distinction. The energy compares these alternatives locally. It is a hand-designed engineering objective, not thermodynamic energy, a learned probability, or a guarantee of globally optimal memory organization.

The controller also applies stability rules. Clearly compatible episodes can be reinforced directly. An isolated strong conflict without recurrence or an explicit change cue is retained as an exception, rather than immediately replacing the schema. Otherwise, a feasible alternative must improve the energy beyond an effective stability margin.

The coefficients and thresholds are hand-set and remain frozen for LongMemEval. A separate 300-question LoCoMo development subset is disjoint from the test question IDs but shares the same histories. The available calibration record does not establish prospective tuning on this subset.

\subsection{Retrieving Original Evidence}
\label{sec:retrieval}
A local router identifies whether a question asks about a current, historical, contextual, or generic state. Retrieval combines direct dense and BM25 search over raw episodes, schema search expanded to support and exception episodes, and representative medoids from active and historical versions. Direct retrieval can recover a detail that a schema route misses, while schema and version routes expose the relevant state and its evidence.

We use all-MiniLM-L6-v2 embeddings and the ms-marco-MiniLM-L6-v2 cross-encoder. Reranking operates on a deduplicated union of up to 56 candidates and returns eight evidence items. Primary MINDSET uses these raw episodes directly. Additional-backbone and LongMemEval experiments apply passage adaptation, preserving source IDs and offsets. Appendix~\ref{app:protocol} distinguishes these evidence budgets from the transition coefficients.

\mindset\ performs no gradient training and makes no provider call during ingestion or retrieval. Its learned components are frozen local encoders. This avoids generative rewriting at write time, but does not make retrieval free: schema search, provenance expansion, and reranking incur measurable latency, evaluated in Section~\ref{sec:efficiency}.

\section{Experimental Setup}
We use dataset subsets instead of entire benchmarks to manage API costs.
\label{sec:setup}
\subsection{Datasets}
\paragraph{LoCoMo.}
We use a frozen 700-question manifest containing 70 questions from each of ten conversations \citep{ref1}. It contains 151 single-hop, 151 temporal, 96 multi-hop, 151 open-domain, and 151 adversarial questions. Answer metrics use all 700 questions; ranking metrics use the 696 with nonempty gold evidence. The released development set contains 300 other question IDs from the same ten histories (Appendix~\ref{app:development}). The cross-dataset evaluation provides a stronger test of transfer beyond these shared histories.

\paragraph{LongMemEval knowledge updates.}
To test the capability most directly targeted by the controller, we evaluate the frozen \mindset\ configuration on all 78 knowledge-update cases in the cleaned LongMemEval-S release \citep{ref3}, without retuning. All six methods use the native task-prompt judge with Gemini-3.1-Flash-Lite substituted for the reference judge. We report judge accuracy, token F1, and selected recall diagnostics for this category.

\subsection{Baselines and Answer Models}
We compare six systems: \mindset, Vector RAG, a recency-truncated Long Context baseline, MemoryBank, LightMem and Nemori. Vector RAG uses normalized MiniLM embeddings over timestamped dialogue turns. Long Context supplies chronological history with deterministic oldest-turn removal beyond a 20,000-byte UTF-8 limit. This cap is part of our protocol; the baseline does not represent unrestricted full-context inference.

MemoryBank uses the shared GPT-OSS model for summary calls in place of its original summarizer. LightMem is a documented reproduction with substitutions for unavailable selector and consolidation components. Nemori is reproduced from its official implementation. These adaptations support comparison and usefulness of our system within a common environment. We do not claim about general leaderboard standing or superiority.
The primary LoCoMo runs use GPT-OSS-120B. Additional complete LoCoMo-700 runs use GLM-4.7 and Gemma-4-31B. Within each run, systems share answer instructions requiring concise, evidence-grounded answers and abstention when evidence is insufficient. The additional runs assess backbone compatibility. Appendix~\ref{app:backbones} summarizes these experiments.

\subsection{Evidence Budgets and Metrics}
Primary LoCoMo uses eight native memory hits, except that MemoryBank applies a passage adapter with at most eight passages and an 8,000-token ceiling. The additional-backbone runs apply this adapter to all retrieval methods. Long Context instead uses its chronological byte cap. These recorded settings differ in representation and, across experiments, evidence adaptation.

LoCoMo answer F1 uses deterministic, category-aware scoring: normalization removes punctuation and articles and applies Porter stemming; multiple references are supported, and adversarial questions reward recognized abstention. We also report Recall@8, Precision@8, mean reciprocal rank (MRR), nDCG@8, and provenance-based evidence recall. Long Context's retrieval columns describe its adapted evidence representation rather than an explicit ranked retriever.

LongMemEval judge accuracy is the proportion of answers accepted by the evaluator; it is distinct from deterministic token F1. We do not compare our LoCoMo token-F1 results directly with published judge-based leaderboard scores.

\section{Results and Analysis}
\label{sec:results}
\subsection{Direct Evaluation of Knowledge Updates}
\label{sec:updates}
\begin{table}[t]
\centering\small
\setlength{\tabcolsep}{5pt}
\begin{tabular}{lrr}
\toprule
Method & Judge accuracy (\%) & Token F1 \\
\midrule
\mindset & \best{69.23} & \best{0.4708} \\
Vector RAG & 64.10 & 0.4312 \\
LightMem & 60.26 & 0.4391 \\
Nemori & 60.26 & 0.3883 \\
MemoryBank & 32.05 & 0.0962 \\
Long Context & 23.08 & 0.0930 \\
\bottomrule
\end{tabular}
\caption{All 78 LongMemEval knowledge-update cases, with MINDSET frozen without retuning. All use the same task-prompt judge protocol. No significance claim is made.}
\label{tab:lme}
\end{table}

Table~\ref{tab:lme} provides the most direct evidence for the proposed update mechanism. \mindset\ achieves 69.23\% judge accuracy and 0.4708 token F1, the highest observed values among the evaluated systems. Vector RAG is the closest baseline in judge accuracy at 64.10\%. These percentages correspond to 54 and 50 accepted answers out of 78, respectively: a four-case difference. The result supports transfer of the frozen controller to an update-focused dataset, while the small subset warrants caution about the size of the advantage.

\subsection{Retrieval and Answering on LoCoMo}
\label{sec:locomo}
\begin{table*}[t]
\centering\small
\setlength{\tabcolsep}{6pt}
\begin{tabular}{lrrrrrr}
\toprule
Method & Answer F1 & Recall@8 & Precision@8 & MRR & nDCG@8 & Evidence recall \\
\midrule
\mindset & \best{0.4280} & \best{0.6280} & \best{0.1056} & \best{0.5572} & \best{0.5453} & 0.6302 \\
LightMem & 0.4024 & 0.4522 & 0.0783 & 0.4365 & 0.4092 & 0.4554 \\
Vector RAG & 0.3807 & 0.3904 & 0.0670 & 0.2675 & 0.2779 & 0.3939 \\
Nemori & 0.3733 & 0.4198 & 0.0682 & 0.2527 & 0.2729 & \best{0.6321} \\
MemoryBank & 0.2975 & 0.0634 & 0.0106 & 0.0438 & 0.0463 & 0.6104 \\
Long Context & 0.2791 & 0.0153 & 0.0036 & 0.0082 & 0.0082 & 0.2190 \\
\bottomrule
\end{tabular}
\caption{LoCoMo-700 with GPT-OSS-120B under our shared protocol. Answer F1 and evidence recall use 700 questions; ranking metrics use 696 questions with annotated evidence. Evidence recall measures expanded provenance coverage, distinct from ranked retrieval. Long Context is recency-truncated and has no explicit ranked retrieval stage. Bold indicates the highest value in each column.}
\label{tab:locomo}
\end{table*}

On LoCoMo-700, \mindset\ obtains 0.4280 answer F1, compared with 0.4024 for LightMem and 0.3807 for Vector RAG (Table~\ref{tab:locomo}). The clearest differences are in ranked retrieval: MRR improves by 0.1207 and nDCG@8 by 0.1361 over LightMem. Nemori has slightly higher provenance-based evidence recall, indicating that recovering evidence somewhere in an expanded memory unit is different from placing useful evidence at effective ranks.

We test the primary LoCoMo differences using two-sided exact paired sign-flip tests over the ten conversation clusters, enumerating all 1,024 sign assignments. Confidence intervals use conversation-level bootstrap resampling. This preserves the dependence among questions from the same history; Holm correction is applied within each metric family.

\begin{figure*}[t]
  \centering
  \resizebox{\textwidth}{!}{
\begin{tikzpicture}[x=1cm,y=1cm,text=figNavy,
  label/.style={font=\sffamily\fontsize{8}{9.5}\selectfont},
  heading/.style={label,font=\sffamily\bfseries\fontsize{10}{12}\selectfont}]
\path[use as bounding box] (-.15,.02) rectangle (16.3,6.05);
\node[heading,anchor=west] at (.15,5.75) {(a) Ranked retrieval};
\node[heading,anchor=west] at (8.35,5.75) {(b) Answer-F1 difference};
\fill[figTeal] (.5,5.02) rectangle (.78,5.22);
\node[label,anchor=west] at (.9,5.12) {MINDSET};
\fill[figBlue!45] (3.45,5.02) rectangle (3.73,5.22);
\node[label,anchor=west] at (3.85,5.12) {LightMem};
\foreach \v in {0,.2,.4,.6} {
  \pgfmathsetmacro{\yy}{.9+4.857143*\v}
  \draw[draw=figBorder,line width=.5pt] (.55,\yy) -- (7.35,\yy);
  \node[label,anchor=east,text=figSlate] at (.4,\yy)
    {\pgfmathprintnumber[fixed,precision=1]{\v}};
}
\draw[draw=figSlate,line width=.7pt] (.55,.9) -- (7.35,.9);
\foreach \x/\ours/\base/\metric in {
  1.65/.6280/.4522/Recall@8,
  4.0/.5572/.4365/MRR,
  6.35/.5453/.4092/nDCG@8} {
  \pgfmathsetmacro{\yo}{.9+4.857143*\ours}
  \pgfmathsetmacro{\yb}{.9+4.857143*\base}
  \fill[figTeal] (\x-.49,.9) rectangle (\x-.04,\yo);
  \fill[figBlue!45] (\x+.04,.9) rectangle (\x+.49,\yb);
  \node[label,anchor=south,font=\sffamily\fontsize{7.4}{9}\selectfont]
    at (\x-.265,\yo+.07)
    {\pgfmathprintnumber[fixed,precision=4,zerofill]{\ours}};
  \node[label,anchor=south,font=\sffamily\fontsize{7.4}{9}\selectfont]
    at (\x+.265,\yb+.07)
    {\pgfmathprintnumber[fixed,precision=4,zerofill]{\base}};
  \node[label] at (\x,.57) {\metric};
}
\node[label,text=figSlate] at (3.95,.15) {Score (higher is better)};
\foreach \tick in {0,.05,.10,.15,.20} {
  \pgfmathsetmacro{\xx}{11.5+20*\tick}
  \draw[draw=figBorder,line width=.5pt] (\xx,1.14) -- (\xx,4.98);
  \draw[draw=figSlate,line width=.6pt] (\xx,1.14) -- (\xx,1.06);
  \node[label,anchor=north,text=figSlate] at (\xx,1)
    {\pgfmathprintnumber[fixed,precision=2,zerofill]{\tick}};
}
\draw[densely dashed,draw=figSlate,line width=.8pt]
  (11.5,1.14) -- (11.5,4.98);
\draw[draw=figSlate,line width=.7pt] (11,1.14) -- (15.75,1.14);
\foreach \y/\name/\mean/\lo/\hi/\col in {
  4.65/LightMem/.0256/-.0118/.0636/figAmber,
  3.9/Vector RAG/.0473/.0178/.0783/figTeal,
  3.15/Nemori/.0547/.0188/.1074/figTeal,
  2.4/MemoryBank/.1305/.0786/.1822/figTeal,
  1.65/Long Context/.1489/.1094/.1857/figTeal} {
  \pgfmathsetmacro{\xm}{11.5+20*\mean}
  \pgfmathsetmacro{\xl}{11.5+20*\lo}
  \pgfmathsetmacro{\xr}{11.5+20*\hi}
  \node[label,anchor=east] at (10.82,\y) {\name};
  \draw[draw=\col,line width=1.25pt] (\xl,\y) -- (\xr,\y);
  \draw[draw=\col,line width=.85pt] (\xl,\y-.09) -- (\xl,\y+.09);
  \draw[draw=\col,line width=.85pt] (\xr,\y-.09) -- (\xr,\y+.09);
  \fill[\col] (\xm,\y) circle (.075);
  \node[label,anchor=south,font=\sffamily\fontsize{7.5}{9}\selectfont]
    at (\xm,\y+.12)
    {$+\pgfmathprintnumber[fixed,precision=4,zerofill]{\mean}$};
}
\node[label,text=figSlate] at (13.35,5.35) {Mean and 95\% cluster-bootstrap interval};
\node[label,text=figSlate] at (13.2,.37) {$\Delta$ answer F1 (MINDSET $-$ baseline)};
\end{tikzpicture}}
  \caption{\textbf{Retrieval gains and answer uncertainty on LoCoMo.} Left: ranked retrieval for \mindset\ and LightMem on 696 evidence-annotated questions; all three comparisons have Holm-adjusted $p=0.00977$. Right: answer-F1 differences against all five baselines on 700 questions, with 95\% conversation-cluster bootstrap intervals. The LightMem interval crosses zero and its adjusted $p=0.24023$; the other four comparisons satisfy adjusted $p<0.05$. Values correspond to Tables~\ref{tab:locomo} and~\ref{tab:full-locomo-statistics}.}
  \label{fig:retrieval-answer}
\end{figure*}
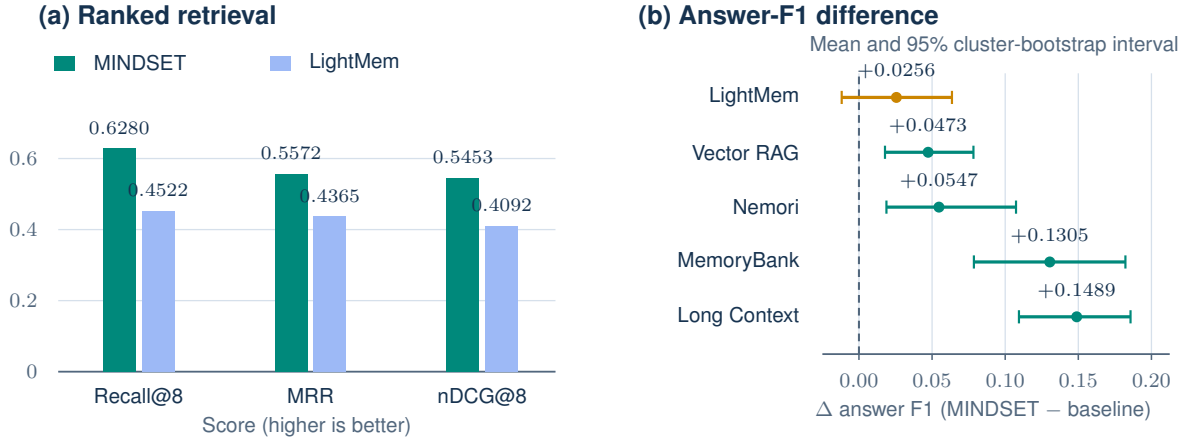

The gains over LightMem in Recall@8, MRR, and nDCG@8 each have Holm-adjusted $p=0.00977$. In contrast, the answer-F1 difference of 0.0256 has a 95\% confidence interval of $[-0.0118,0.0636]$ and adjusted $p=0.24023$. Thus, \mindset\ has the highest observed answer F1, but its advantage over LightMem is not statistically confirmed. Answer-F1 differences against the other four baselines meet the adjusted $p<0.05$ criterion. Figure~\ref{fig:retrieval-answer} distinguishes these findings; Appendix~\ref{app:statistics} gives the full tests. These results apply to primary LoCoMo, not LongMemEval.

\subsection{Which Components Matter?}
\label{sec:ablation}
\begin{table}[t]
\centering\small
\setlength{\tabcolsep}{5pt}
\begin{tabular}{lrr}
\toprule
Variant & Answer F1 & Recall@8 \\
\midrule
Full \mindset & \best{0.4280} & 0.6280 \\
Nearest-schema merge & 0.4080 & \best{0.6296} \\
No contextual term & 0.4194 & 0.6280 \\
No temporal term & 0.4237 & 0.6266 \\
No hysteresis & 0.4237 & 0.6280 \\
No fragmentation penalty & 0.4043 & 0.6256 \\
No provenance expansion & 0.4274 & 0.6158 \\
\bottomrule
\end{tabular}
\caption{Six ablation conditions on LoCoMo-700 with GPT-OSS-120B. Intervention definitions appear in Appendix~\ref{app:ablations}.}
\label{tab:ablation}
\end{table}

Replacing transition selection with nearest-schema merging reduces answer F1 by 0.0200, even though Recall@8 is slightly higher. Removing the fragmentation penalty produces the largest observed decrease, 0.0237 (Table~\ref{tab:ablation}). Both observations support the interpretation that useful organization matters beyond retrieving more gold evidence. Removing contextual, temporal, or stability components has smaller individual effects. Removing provenance expansion lowers Recall@8 but changes answer F1 only slightly. These results do not justify claiming that every component is essential. They identify controlled fragmentation and transition assignment as the strongest observed contributors within this evaluation.

\subsection{Compatibility Across Answer Models}
\label{sec:backbones}
Additional full LoCoMo-700 runs give \mindset\ answer F1 of 0.4080 with GLM-4.7, compared with 0.4005 for LightMem. With Gemma-4-31B, LightMem is marginally higher at 0.3308 versus 0.3306 for \mindset. \mindset\ leads the reported ranking metrics in both additional runs. Unlike primary MINDSET, these runs retrieve twenty initial hits and apply passage adaptation before answering. They demonstrate backbone compatibility but do not isolate a generator-only change.

\subsection{Runtime and Evidence Size}
\label{sec:efficiency}
\begin{table}[t]
\centering\small
\setlength{\tabcolsep}{4pt}
\begin{tabular}{lrrr}
\toprule
Method & Ingest (s) & Retrieve (s) & Packet (KB) \\
\midrule
\mindset & 74.6 & 33.1 & 9.2 \\
Vector RAG & 65.6 & 4.2 & 8.8 \\
LightMem & 101.1 & 10.5 & 9.1 \\
Nemori & 103.9 & 9.4 & 25.2 \\
MemoryBank & 4.0 & 0.13 & 8.0 \\
Long Context & n/a & --- & 19.2 \\
\bottomrule
\end{tabular}
\caption{Mean ingestion time per case, retrieval time per query, and evidence size (decimal KB) on LongMemEval. Long Context has no explicit construction or ranked-retrieval stage. Measurements were taken on a Windows CPU host and are not hardware-normalized comparisons.}
\label{tab:runtime}
\end{table}

\mindset's ingestion averages 74.6 seconds per case, below LightMem and Nemori but above Vector RAG and MemoryBank (Table~\ref{tab:runtime}). Its retrieval is the slowest: 33.1 seconds per query, versus 10.5 for LightMem, 9.4 for Nemori, and 4.2 for Vector RAG. Measurements were collected on a Windows host with four logical CPUs and include candidate search, batched cross-encoding, and passage adaptation. The cross-encoder is loaded once rather than per query.

The average evidence packet is 9.2 KB, compared with 25.2 KB for Nemori and 19.2 KB for Long Context, and close to LightMem and Vector RAG. Packet size is a serialized evidence-size measure, not model-token consumption or end-to-end latency. The primary LoCoMo run logs 403,978 provider tokens for \mindset\ and 4,310,788 for the capped Long Context baseline.

\section{Discussion}
The experiments support a specific view of conversational memory: preserving evidence and organizing its changing interpretation are separate responsibilities. \mindset\ uses versioned schemas to expose current and historical states without asking a language model to rewrite the record during every update. Its energy score makes the trade-off between incompatible merging and unnecessary branching explicit. The evidence is strongest for improved LoCoMo retrieval ranking and promising update-focused transfer. The results motivate further analysis of which conflicts should be held as exceptions. They also reveal a practical bottleneck: compact evidence packets do not compensate automatically for slow retrieval.

\section{Conclusion}
\mindset\ manages changing conversational knowledge through four explicit transitions, a local energy score, and stability rules that preserve original evidence. The design improves LoCoMo retrieval ranking and transfers to knowledge updates without retuning. Answer gains vary across baselines and backbones, and retrieval latency remains substantial. These findings support explicit transition control as a useful component of conversational memory.

\section*{Limitations}
Our findings are bounded by the evaluation protocol. LoCoMo does not provide a conventional judge-based leaderboard comparison and uses a subset of 700 questions only rather than the entire benchmark. Development and test question IDs are disjoint, but the underlying conversation histories need not be. The controller uses hand-set coefficients, lexical conflict signals and a limited attribute schema rather than scientifically proven optimal values.

\FloatBarrier

\bibliography{references}
\FloatBarrier
\clearpage
\appendix
\section{Reproducibility and Evaluation Protocol}
\label{app:protocol}
Transition coefficients determine the update policy, whereas evaluation budgets constrain the evidence and model requests. This appendix specifies both so that comparisons can be interpreted independently of software organization.

\paragraph{Local components.}
Frozen all-MiniLM-L6-v2 embeddings and the ms-marco-MiniLM-L6-v2 cross-encoder are used with seed zero. Ingestion searches up to six nearby active schemas. Retrieval combines up to 24 direct episodes, 24 schema-provenance episodes, and four medoids from each active/historical status, yielding at most 56 candidates before deduplication. Cross-encoding uses batches of 32.

\paragraph{Shared evidence adapter.}
Where enabled, up to 20 memories are segmented into 768-token passages with 96-token overlap, rescored with the local embedding model, and greedily selected with at most eight passages, 8,000 tokens, and three passages per source memory. Passage records retain source IDs and offsets. This adapter is enabled only for MemoryBank in primary LoCoMo, for retrieval methods in additional-backbone runs, and for MINDSET, Vector RAG, LightMem and Nemori in LongMemEval. LongMemEval MemoryBank uses compact memories without that adapter. Long Context removes oldest turns at a 20,000-byte UTF-8 cap.

\paragraph{Answer generation.}
Primary GPT-OSS-120B runs use Cerebras, temperature zero, structured JSON validation, and a 4,096-token output reservation. Additional GLM-4.7 and Gemma-4-31B runs use one question per request, a 1,024-token cap, and a 160,000-byte prompt guard.

\paragraph{Answer instructions.}
Answer models receive a concise, evidence-grounded instruction with evidence isolated for each question. Names, dates, attributes, and list members must be supported by that question's evidence; otherwise the required response is ``No information available.'' Structured responses preserve question identities, and bounded retries handle missing or malformed answers.

\section{Development Set and Calibration}
\label{app:development}
The 300-question LoCoMo development set excludes all primary LoCoMo-700 question IDs. It contains 75 questions from each available category (1, 2, 4, and 5). Category 3 has no remaining examples because it is fully represented in the primary subset. Within each category, questions are ordered by a deterministic SHA-256 key, the first 75 are retained, and the selection is globally ordered by the same key. Compatibility coefficients, energy weights, thresholds, and stability margins are hand-set. LongMemEval uses the same controller settings without retuning.

\paragraph{Practical tuning guide.}
Domain adaptation should begin with the feature extractor and the distinction between single- and multi-valued attributes. Development examples should cover genuine updates, contextual coexistence, unrelated topics, and isolated conflicts. Incorrect merges and excessive small schemas indicate whether compatibility and feasibility thresholds need adjustment before the energy penalties. Fragmentation controls unnecessary branching, whereas the exception rule may delay legitimate corrections. Parameters should be frozen before evaluation.

\section{Transition Procedure and Illustrative Examples}
\label{app:algorithm}
Episodes are processed chronologically, and their original text remains unchanged. The transition policy follows five ordered stages:
\begin{enumerate}[leftmargin=*,itemsep=2pt,topsep=3pt]
  \item If no active schema exists, create a new schema.
  \item Otherwise, choose the most compatible nearby schema and assess conflict, temporal change, and context.
  \item Create a new schema for a clear topic mismatch; reinforce a clearly compatible state; retain an isolated strong conflict as an exception.
  \item For remaining cases, compare feasible alternatives with reinforcement. Select an alternative only when its energy advantage exceeds the effective stability margin.
  \item Apply the transition, preserving original evidence and linking successive versions when supersession occurs.
\end{enumerate}
Reinforcement can therefore add either supporting evidence or an unresolved exception.


\paragraph{Observed transitions.}
The observed controller decisions include reinforcement of compatible evidence, supersession when a previously stored state changes, creation of a schema for a new topic, and contextual splitting when apparently conflicting states refer to different circumstances.

\subsection{Functional Specification of the Frozen Controller}
\label{app:functional}

The controller compares memory transitions using a local energy objective, subject to feasibility and stability guards. This objective is not thermodynamic energy. The following parameters remain fixed for full-controller evaluations. Ablations modify only their designated components.

\paragraph{Compatibility and contradiction.}
Let $\mathbf v$ be the incoming episode embedding, $\mathbf c$ a
schema centroid, and $s=\max(0,\cos(\mathbf c,\mathbf v))$.
Entity, role, and context overlaps $e,r,c$ use
\[
 O(B,Y)=\frac{\sum_{y\in Y}B(y)}{\sum_y B(y)},
\]
where $B$ contains schema feature counts and $Y$ is the incoming
set; the overlap is $1/2$ when either is empty. Let $b$ indicate
that a non-discourse incoming attribute slot matches the schema's
topic suffix. Compatibility is clipped to $[0,1]$:
\[
 \kappa=0.52s+0.18e+0.12r+0.10c+0.08b.
\]
Candidate search prefers schemas with the same speaker--slot
topic, falling back to all active schemas. It retains the 64 most
recently created schemas in this pool and selects up to six by
embedding similarity. The target maximizes
$(\kappa,s,\mathrm{ID})$ lexicographically within this candidate set.

For shared single-valued attributes $j\in\mathcal S$,
\[
 p=\max_{j\in\mathcal S}
 \left[1-\max_{u\in V_j,\,v\in V'_j}J(u,v)\right],
\]
where $V_j,V'_j$ are stored and incoming values. $J$ is Jaccard
similarity over lowercased alphanumeric tokens after stop-word
removal, with exact normalized-string equality used if either
token set is empty. Set $p=0$ when $\mathcal S$ is empty.
Preference, dislike, plan, possession, experience, and ability
slots are treated as multi-valued and excluded from this test.

\paragraph{Temporal and contextual potentials.}
Let $M$ count exception episodes matching any incoming attribute
value in a shared slot at $J\geq0.8$, and set
$\rho=\min(1,M/2)$. Let $\ell$ indicate that the incoming timestamp
is at least the latest supporting episode's timestamp, and let
$z$ indicate a lexical change cue. Unparseable timestamps map to
zero. The potentials are
\[
\begin{aligned}
 T &=p\min(1,0.35c+0.20\ell\\
   &\hspace{15mm}+0.40z+0.25\rho),\\
 X &=\max(p,0.65b)\min(1,0.45(1-c)\\
   &\hspace{15mm}+0.20(1-r)+0.35(1-s)).
\end{aligned}
\]
$T$ supports temporal updates, whereas $X$ supports contextual
coexistence.

\paragraph{Distortion and fragmentation.}
For $n$ supporting episodes and stored recursive distortion
$\bar d$, define
\[
\begin{aligned}
 \mathbf c'&=\operatorname{unit}(n\mathbf c+\mathbf v),\\
 d_c&=1-\cos(\mathbf c,\mathbf c'),\\
 d_e&=1-\cos(\mathbf v,\mathbf c'),\\
 \widehat D&=\frac{n(\bar d+d_c)+d_e}{n+1}.
\end{aligned}
\]
Zero vectors remain zero under normalization, and their cosine
similarity is defined as zero. Let $J_S$ be the mean impurity
$1-\max_y B(y)/\sum_y B(y)$ over the current nonempty attribute,
role, and context histograms, or zero if none exist.

With $A$ active schemas, $Q$ undersized active schemas,
$m=2$, $\eta=0.75$, and stored-episode count $N_{\rm ep}$,
\[
\begin{aligned}
 F_R&=\frac{
 A+\eta\max\{0,Q-\mathbf{1}_{\{n<m\leq n+1\}}\}
 }{\sqrt{N_{\rm ep}+1}},\\
 F_U&=\frac{A+\eta Q}{\sqrt{N_{\rm ep}+1}},\\
 F_B&=\frac{A+1+\eta(Q+1)}{\sqrt{N_{\rm ep}+1}}.
\end{aligned}
\]
$N_{\rm ep}$ includes the incoming episode, which is stored before
transition selection. These are local complexity proxies:
supersession leaves the active-count penalty unchanged and does
not recompute the successor's undersized status.

\paragraph{Action energies.}
Write $R,U,S$ for reinforcement, supersession, and splitting,
and let $[x]_+=\max(x,0)$. The action energies are
\[
\begin{aligned}
 E_R&=w_D\widehat D+w_Cp+w_H(d_c+0.35p)\\
    &\quad+w_FF_R+w_I[J_S+0.5\max(T,X)],\\
 E_U&=w_FF_U+w_I[X-T]_+,\\
 E_S&=w_FF_B+w_I[T-X]_+,\\
 E_{\mathrm{NEW}}&=w_FF_B+w_I\kappa.
\end{aligned}
\]
The weights for distortion, contradiction, historical damage,
fragmentation, and inconsistency are
\[
 (w_D,w_C,w_H,w_F,w_I)=(1,1.25,0.75,0.22,1).
\]
Candidate costs are evaluated before the final support/exception
assignment; they are not a global recomputation of memory quality.

\paragraph{Feasibility and switching.}
If no active schema exists, the controller creates one.
Otherwise, it first selects NEW when
$\kappa<0.43$, $b=0$, and $s<0.35$.
For remaining cases, the feasible set starts with
$\{R,\mathrm{NEW}\}$, admits $U$ when $p\geq0.35$ and $S$ when
$X\geq0.12$, and excludes NEW when $\kappa\geq0.43$.
A compatible episode reinforces directly when
$p<0.15$, $X<0.12$, and $\kappa\geq0.58$.
An isolated conflict is held by reinforcement when
$p\geq0.60$, $\rho=0$, and $z=0$.

After these guards, let $a^*$ be the lowest-energy feasible
non-reinforcing action. Ties are resolved in the order
NEW, SPLIT, SUPERSEDE. Define
\[
\begin{aligned}
 g&=E_R-E_{a^*},\\
 h&=0.18(1-0.6\rho)b_T,\\
 b_T&=
 \begin{cases}
 0.55,&T>X\ \text{and}\ T>0.45,\\
 1,&\text{otherwise}.
 \end{cases}
\end{aligned}
\]
For the full controller, switching then occurs exactly when
\[
 g>\min(h,0.06).
\]
Otherwise, or if no non-reinforcing action is feasible, the
controller reinforces.

A reinforcing episode with $p\geq0.45$ is retained as an
exception without updating the centroid, feature histograms,
or recursive distortion. Other reinforcing episodes update
support. Supersession closes the prior version and creates
a linked successor; splitting creates a coexisting active
schema without a stored parent link. Original episodes are
never rewritten.

\section{Additional Answer-Backbone Evaluation}
\label{app:backbones}

We evaluate all six methods on the same 700 LoCoMo questions using GLM-4.7 and Gemma-4-31B. Answer F1 and evidence recall use all 700 questions; ranking metrics use the 696 questions with nonempty gold evidence. R@8 and P@8 abbreviate Recall@8 and Precision@8, respectively.

These experiments use one question per answer request and a
1,024-token output limit. Retrieval methods obtain twenty
initial memory hits and apply passage selection, retaining
at most eight passages within an 8,000-token budget.
Long Context instead retains chronological history within
a 20,000-byte limit. Unlike primary MINDSET, which supplies
eight raw episodes, these runs use adapted evidence.

\subsection{GLM-4.7 Results}
\label{app:glm-results}

\par\vspace{8pt}
\noindent
\begin{minipage}{\linewidth}
\centering
\small
\setlength{\tabcolsep}{4pt}
\begin{tabular}{@{}lrrr@{}}
\toprule
Method & \shortstack{Answer\\F1}
       & \shortstack{Evidence\\recall}
       & \shortstack{Logged\\tokens} \\
\midrule
MINDSET      & \textbf{0.4080} & 0.4326 & 507,238 \\
LightMem     & 0.4005 & 0.4164 & 708,450 \\
Vector RAG   & 0.3815 & 0.3939 & 609,491 \\
Nemori       & 0.3961 & 0.4661 & 955,224 \\
MemoryBank   & 0.2659 & \textbf{0.8952} & 831,244 \\
Long Context & 0.2975 & 0.2190 & 4,360,926 \\
\bottomrule
\end{tabular}
\captionof{table}{GLM-4.7 answer quality, expanded evidence
coverage, and recorded provider-token usage on LoCoMo-700.
Token totals are subject to the accounting qualifications below.}
\label{tab:glm-answer}
\end{minipage}
\par\vspace{8pt}

\noindent
\begin{minipage}{\linewidth}
\centering
\small
\setlength{\tabcolsep}{4pt}
\begin{tabular}{@{}lrrrr@{}}
\toprule
Method & R@8 & P@8 & MRR & nDCG@8 \\
\midrule
MINDSET
 & \textbf{0.4293} & \textbf{0.0745}
 & \textbf{0.2806} & \textbf{0.2957} \\
LightMem
 & 0.4130 & 0.0718 & 0.2744 & 0.2875 \\
Vector RAG
 & 0.3904 & 0.0670 & 0.2675 & 0.2779 \\
Nemori
 & 0.3085 & 0.0510 & 0.1794 & 0.1937 \\
MemoryBank
 & 0.0671 & 0.0115 & 0.0501 & 0.0494 \\
Long Context
 & 0.0153 & 0.0036 & 0.0082 & 0.0082 \\
\bottomrule
\end{tabular}
\captionof{table}{GLM-4.7 ranking metrics on 696
evidence-annotated questions. Long Context's scores describe
its adapted evidence representation rather than an explicit
ranked retriever.}
\label{tab:glm-retrieval}
\end{minipage}
\par\vspace{8pt}

MINDSET obtains the highest observed answer F1 by a modest margin but leads all four ranking metrics
(Tables~\ref{tab:glm-answer}--\ref{tab:glm-retrieval}).
MemoryBank's substantially higher expanded evidence recall
does not translate into higher answer F1 or ranked retrieval quality, illustrating the distinction between provenance coverage and effective evidence selection. Further analysis on the token usage and ranking metrics has been done in details after the Gemma-4-31B results section below.  

\subsection{Gemma-4-31B Results}
\label{app:gemma-results}

\par\vspace{8pt}
\noindent
\begin{minipage}{\linewidth}
\centering
\small
\setlength{\tabcolsep}{4pt}
\begin{tabular}{@{}lrrr@{}}
\toprule
Method & \shortstack{Answer\\F1}
       & \shortstack{Evidence\\recall}
       & \shortstack{Logged\\tokens} \\
\midrule
MINDSET      & 0.3306 & 0.4326 & 768,173 \\
LightMem     & \textbf{0.3308} & 0.4164 & 825,896 \\
Vector RAG   & 0.3226 & 0.3939 & 705,867 \\
Nemori       & 0.3295 & 0.4661 & 1,440,066 \\
MemoryBank   & 0.2582 & \textbf{0.8657} & 979,396 \\
Long Context & 0.2739 & 0.2190 & 5,000,182 \\
\bottomrule
\end{tabular}
\captionof{table}{Gemma-4-31B answer quality, expanded
evidence coverage, and recorded provider-token usage on
LoCoMo-700.}
\label{tab:gemma-answer}
\end{minipage}
\par\vspace{8pt}

\noindent
\begin{minipage}{\linewidth}
\centering
\small
\setlength{\tabcolsep}{4pt}
\begin{tabular}{@{}lrrrr@{}}
\toprule
Method & R@8 & P@8 & MRR & nDCG@8 \\
\midrule
MINDSET
 & \textbf{0.4293} & \textbf{0.0745}
 & \textbf{0.2806} & \textbf{0.2957} \\
LightMem
 & 0.4130 & 0.0718 & 0.2744 & 0.2875 \\
Vector RAG
 & 0.3904 & 0.0670 & 0.2675 & 0.2779 \\
Nemori
 & 0.3085 & 0.0510 & 0.1794 & 0.1937 \\
MemoryBank
 & 0.0678 & 0.0115 & 0.0453 & 0.0488 \\
Long Context
 & 0.0153 & 0.0036 & 0.0082 & 0.0082 \\
\bottomrule
\end{tabular}
\captionof{table}{Gemma-4-31B ranking metrics on 696
evidence-annotated questions. Long Context's retrieval scores
have the same diagnostic interpretation as in
Table~\ref{tab:glm-retrieval}.}
\label{tab:gemma-retrieval}
\end{minipage}
\par\vspace{8pt}

LightMem obtains the highest observed answer F1, although
its advantage over MINDSET is approximately 0.0002
(Table~\ref{tab:gemma-answer}). Nemori is also close at
0.3295. MINDSET nevertheless leads all four ranking metrics
(Table~\ref{tab:gemma-retrieval}). These results support
compatibility with the additional answer backbones, but not
a uniform answer-F1 advantage among the closely scoring methods.

\paragraph{Interpretation and token accounting.}
MINDSET's identical ranking scores across these two backbones should not be interpreted as independent retrieval evidence. Its Recall@8 is 0.4293 under the additional-run protocol, compared with 0.6280 in the primary experiment. Because evidence adaptation and retrieval implementation differ, this change cannot be attributed solely to the answer model or solely to passage selection.

Logged tokens are the provider-recorded input-plus-output
totals for these answer runs. The GLM MINDSET and Nemori runs contain 163 and 177 cache hits, respectively; the Gemma runs contain none. Consequently, the totals document the reported executions but do not establish an intrinsic token-efficiency ranking. Overall, the additional experiments show that MINDSET remains competitive across answer models, with consistently strong ranked retrieval under this protocol and backbone-dependent answer gains.

\section{LoCoMo Statistical Results}
\label{app:statistics}
The primary LoCoMo analysis compares MINDSET against the evaluated baselines, with the main text highlighting LightMem and Nemori. Tests use paired scores clustered by conversation, with Holm correction for multiple comparisons within each metric. Ranking metrics exclude questions without gold evidence, while answer F1 uses the full evaluation set. Inference uses unrounded per-question scores.

Answer-F1 gains are supported against several baselines, but the LightMem comparison is not statistically significant under the reported procedure. Ranking gains against LightMem and Nemori are more consistent. These findings are specific to LoCoMo; no corresponding significance claim is made for LongMemEval. Exact statistical outputs and reproduction settings can be reproduced from the accompanying code and configuration release.

\begin{strip}
\centering
\scriptsize

\resizebox{\textwidth}{!}{%
\begin{tabular}{llrrrrr}
\toprule
Baseline & Metric & Mean $\Delta$ & 95\% CI
& W--T--L & Exact $p$ & Holm $p$ \\
\midrule
Vector RAG & Answer F1 & +0.0473 & [+0.0178,+0.0783] & 202--367--131 & 0.02148 & 0.04297 \\
Long Context & Answer F1 & +0.1489 & [+0.1094,+0.1857] & 295--313--92 & 0.00195 & 0.00977 \\
Nemori & Answer F1 & +0.0547 & [+0.0188,+0.1074] & 217--325--158 & 0.00977 & 0.02930 \\
LightMem & Answer F1 & +0.0256 & [-0.0118,+0.0636] & 175--380--145 & 0.24023 & 0.24023 \\
MemoryBank & Answer F1 & +0.1305 & [+0.0786,+0.1822] & 284--318--98 & 0.00391 & 0.01562 \\
\midrule
Vector RAG & Recall@8 & +0.2377 & [+0.1970,+0.2815] & 220--453--23 & 0.00195 & 0.00977 \\
Long Context & Recall@8 & +0.6127 & [+0.5745,+0.6489] & 485--202--9 & 0.00195 & 0.00977 \\
Nemori & Recall@8 & +0.2082 & [+0.1788,+0.2363] & 243--384--69 & 0.00195 & 0.00977 \\
LightMem & Recall@8 & +0.1758 & [+0.1345,+0.2227] & 173--488--35 & 0.00195 & 0.00977 \\
MemoryBank & Recall@8 & +0.5646 & [+0.5189,+0.6030] & 460--221--15 & 0.00195 & 0.00977 \\
\midrule
Vector RAG & Precision@8 & +0.0386 & [+0.0324,+0.0451] & 220--453--23 & 0.00195 & 0.00977 \\
Long Context & Precision@8 & +0.1020 & [+0.0946,+0.1088] & 485--202--9 & 0.00195 & 0.00977 \\
Nemori & Precision@8 & +0.0374 & [+0.0301,+0.0437] & 243--384--69 & 0.00195 & 0.00977 \\
LightMem & Precision@8 & +0.0273 & [+0.0205,+0.0350] & 173--488--35 & 0.00195 & 0.00977 \\
MemoryBank & Precision@8 & +0.0950 & [+0.0899,+0.0996] & 460--221--15 & 0.00195 & 0.00977 \\
\midrule
Vector RAG & MRR & +0.2897 & [+0.2594,+0.3218] & 348--299--49 & 0.00195 & 0.00977 \\
Long Context & MRR & +0.5491 & [+0.5211,+0.5796] & 492--193--11 & 0.00195 & 0.00977 \\
Nemori & MRR & +0.3045 & [+0.2725,+0.3327] & 368--238--90 & 0.00195 & 0.00977 \\
LightMem & MRR & +0.1207 & [+0.0915,+0.1500] & 155--492--49 & 0.00195 & 0.00977 \\
MemoryBank & MRR & +0.5134 & [+0.4823,+0.5439] & 475--211--10 & 0.00195 & 0.00977 \\
\midrule
Vector RAG & nDCG@8 & +0.2674 & [+0.2356,+0.3023] & 364--281--51 & 0.00195 & 0.00977 \\
Long Context & nDCG@8 & +0.5371 & [+0.5067,+0.5674] & 492--193--11 & 0.00195 & 0.00977 \\
Nemori & nDCG@8 & +0.2724 & [+0.2517,+0.2944] & 383--216--97 & 0.00195 & 0.00977 \\
LightMem & nDCG@8 & +0.1361 & [+0.1029,+0.1710] & 183--454--59 & 0.00195 & 0.00977 \\
MemoryBank & nDCG@8 & +0.4990 & [+0.4647,+0.5312] & 474--205--17 & 0.00195 & 0.00977 \\
\midrule
Vector RAG & Evidence recall & +0.2363 & [+0.1959,+0.2797] & 220--457--23 & 0.00195 & 0.00977 \\
Long Context & Evidence recall & +0.4111 & [+0.3673,+0.4575] & 374--264--62 & 0.00195 & 0.00977 \\
Nemori & Evidence recall & -0.0019 & [-0.0303,+0.0272] & 131--426--143 & 0.91602 & 1.00000 \\
LightMem & Evidence recall & +0.1748 & [+0.1343,+0.2202] & 173--492--35 & 0.00195 & 0.00977 \\
MemoryBank & Evidence recall & +0.0198 & [-0.0882,+0.1254] & 190--317--193 & 0.73242 & 1.00000 \\
\bottomrule
\end{tabular}%
}

\captionof{table}{LoCoMo cluster statistics. Ranking metrics use 696 annotated questions; answer F1 and evidence recall use 700.}
\label{tab:full-locomo-statistics}

\end{strip}

\section{Ablation Definitions}
\label{app:ablations}
Nearest-schema merge reinforces the compatibility-selected target after initialization. Contextual and temporal ablations remove their respective contributions from the transition energy while retaining the original feasibility signals. No-hysteresis removes margin stabilization and isolated-conflict holding. No-fragmentation removes the fragmentation penalty. No-provenance-expansion omits schema and version candidates from the active retrieval path. The released configuration records the exact settings used for each comparison.

\section{Local Sensitivity and Memory Structure}
\label{app:sensitivity}
The frozen LoCoMo run produces active and historical schemas while retaining the provenance of supporting and exceptional episodes. These are memory-structure observations, not additional evaluation questions.

The sensitivity analysis varies individual parameters near the frozen configuration while holding the evaluation manifest and retrieval setup fixed, without additional answer-generation or judging calls. Retrieval metrics and memory structure change only slightly across the tested neighborhood. This supports a limited claim of local retrieval and structural stability, not answer robustness or global optimality. The effective transition boundary is determined by the stability margin specified in the transition policy.

\section{LoCoMo Category Breakdown}
\label{app:categories}
Table~\ref{tab:locomo-categories} breaks down LoCoMo results by question type. MINDSET performs best among the compared methods on single-hop, temporal, multi-hop, and open-domain questions, but not adversarial abstention. These comparisons are descriptive rather than additional significance tests.

\begin{strip}
\centering
\small
\begin{tabular}{lrrrrr}
\toprule
Method & Single-hop & Temporal & Multi-hop & Open-domain & Adversarial \\
\midrule
MINDSET & 0.3068 & 0.3902 & 0.2173 & 0.4205 & 0.7285 \\
Vector RAG & 0.2662 & 0.3368 & 0.1659 & 0.2817 & 0.7748 \\
Long Context & 0.1237 & 0.0886 & 0.0828 & 0.1214 & 0.9073 \\
Nemori & 0.2853 & 0.3219 & 0.2026 & 0.3852 & 0.6093 \\
LightMem & 0.2876 & 0.3608 & 0.2092 & 0.2895 & 0.7947 \\
MemoryBank & 0.1515 & 0.1419 & 0.1214 & 0.0683 & 0.9404 \\
\bottomrule
\end{tabular}

\captionof{table}{Category-specific answer scores
on primary LoCoMo-700: 151 single-hop,
151 temporal, 96 multi-hop, 151 open-domain,
and 151 adversarial questions. Evaluation is deterministic and category-aware; adversarial scores reward benchmark-recognized abstention.}
\label{tab:locomo-categories}

\end{strip}

\section{Provider-Token Accounting}
\label{app:tokens}
Provider-recorded token usage varies considerably across methods. Exact cache hits contribute no additional provider tokens in the recorded ledger, while some baselines include memory construction as well as answer generation. The runs also differ in batching and processing strategy, so recorded totals should not be treated as standardized uncached per-query costs. MINDSET's local computation is not represented by provider-token totals. LongMemEval packet sizes are accounted for separately. Exact usage ledgers and experimental settings can be provided with the code release.


\end{document}